# TIME SERIES FORECASTING USING NEURAL NETWORKS

**BOGDAN OANCEA**[*]
**ŞTEFAN CRISTIAN CIUCU**[**]

**Abstract**
*Recent studies have shown the classification and prediction power of the Neural Networks. It has been demonstrated that a NN can approximate any continuous function. Neural networks have been successfully used for forecasting of financial data series. The classical methods used for time series prediction like Box-Jenkins or ARIMA assumes that there is a linear relationship between inputs and outputs. Neural Networks have the advantage that can approximate nonlinear functions. In this paper we compared the performances of different feed forward and recurrent neural networks and training algorithms for predicting the exchange rate EUR/RON and USD/RON. We used data series with daily exchange rates starting from 2005 until 2013.*

## Introduction

Classical statistical and econometric models used for forecasting in the field of financial time series fails to efficiently handle uncertainty nature of foreign exchange data series. One of the largest and more volatile financial market is the foreign exchange market, exchange rates being among the most used and important economic indices. Forecasting the exchange rates is a difficult problem from both theoretical and practical point of view because the exchange rates are influenced by many economic and political factors. During the time, many statistical and econometric models have been developed by researchers for the purpose of forecasting exchange rates but this problem remains one of the major challenges in the field of forecasting methods.

Recent studies have shown the classification and prediction power of the Artificial Neural Networks. It has been demonstrated that a neural network can approximate any continuous function. Neural networks have been successfully used for forecasting of financial data series. The classical methods used for time series prediction like Box-Jenkins, ARMA or ARIMA assumes that there is a linear relationship between inputs and outputs. Neural Networks have the advantage that can approximate any nonlinear functions without any apriori information about the properties of the data series.

In this paper we compared the performances of different feed forward and recurrent neural networks and training algorithms for predicting the exchange rate EUR/RON and USD/RON. We used data series with daily exchange rates starting from 2005 until 2013 provided by the National Bank of Romania. We developed two different kinds of neural networks, a feed forward network and a recurrent network and compared their performances for foreign exchange rate prediction. The feed forward network was trained by the classical backpropagation method, by the resilient backpropagation and one of its' versions, while the recurrent network was trained using a multistream approach based on the Extended Kalman Filter.

## Related work

In (Kondratenko, 2003) the authors presents a forecasting method for exchange rates between American Dollar and Japanese Yen, Swiss Frank, British Pound and EURO using a recurrent neural network. Before the data series were presented to the neural network some preprocessing operations have been done based on normalization, calculation of Hurst exponent, Kolmogorov-Smirnov test in order to remove the possible correlations. An Elman-Jordan recurrent network has been used to forecast the one-day ahead value of moving average of returns with the window equal to 5 observations. The number of hidden neurons was chosen equal to 100 with a linear activation of the input layer and logistic activation of the hidden and output layer. The neural network predicted the increments sign with a high probability – approximately 80%.

---

[*] Professor, PhD, "Nicolae Titulescu" University of Bucharest (email: bogdan.oancea@gmail.com).
[**] IT Director, "Nicolae Titulescu" University (email: stefanciucu@yahoo.com).

(Chen, 2003) compared the performance of a Probabilistic Neural Network with a GMM-Kalman Filter and random walk approach for predicting the direction of return on market index of the Taiwan Stock Exchange. They reached to the conclusion that PNN has a stronger forecasting power than both the GMM–Kalman filter and the random walk models because PNN has a better capability to identify erroneous data and outliers and it also doesn't require any apriori information about the underlying probability density functions of the data.

A comparison between a neural network and a Hidden Markov Model used for foreign exchange forecasting is also given in (Philip 2011). The results of the study show that while the Hidden Markov Model achieved an accuracy of 69.9% the neural network had an accuracy of 81.2%.

A hibrid model ARIMA-PNN is presented in (Khashei, 2012). The values estimated with the ARIMA model are changed based on the trend of the ARIMA residuals detected by a PNN and optimum step length obtained a mathematical programming model.

Data transformation in order to improve the accuracy of the forecast is used in (Proietti, 2013). The authors considered the Box-Cox power transformation and showed the forecasts are improve significantly compared to the untransformed data at the one-step-ahead horizon.

(Man-Chung, 2000) used a conjugate gradient learning algorithm with a restart procedure to improve the convergence. To further improve the convergence the authors also used a multiple linear regression for weight initialization instead of the classical random initialization. They used the network to predict the daily trading data of some companies from Shanghai Stock Exchange.

**Methodology**

In this paper we analyze the exchange rates for two currencies EUR and USD between 03.01.2005 and 19.02.2013. The data series are obtained from the National Bank of Romania. Figure 1 and 2 shows the corresponding exchange rates EUR/RON and USD/RON. It can be seen from these charts that the data are heavily jagged. To improve the accuracy of the forecasting it is better to remove the correlations between the inputs and made them statistically independent. Thus, we studied not the exchange rates but the logarithmic returns given by the following formula:

$$R_n = \frac{\ln E_n}{\ln E_{n-1}}$$

More, in order to increase the learning rate additional data preprocessing that smoothed the data distribution was performed before training the network. We normalized the data using a logistic function according to the following formula:

$$\tilde{R}_n = 1/(1 + \exp(\frac{R_n - \bar{R}}{\sigma}))$$

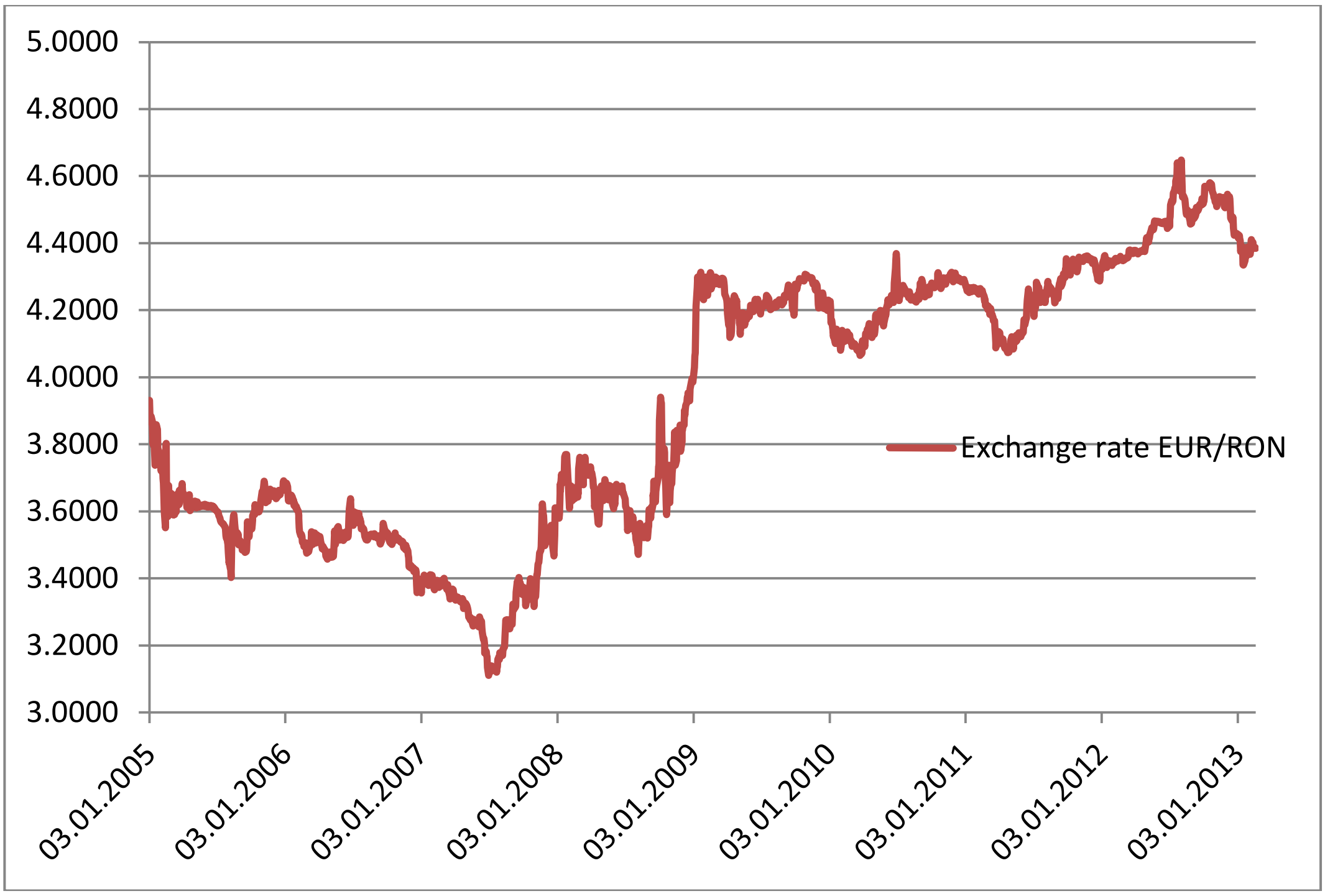

Figure 1. The exchange rate EUR/RON

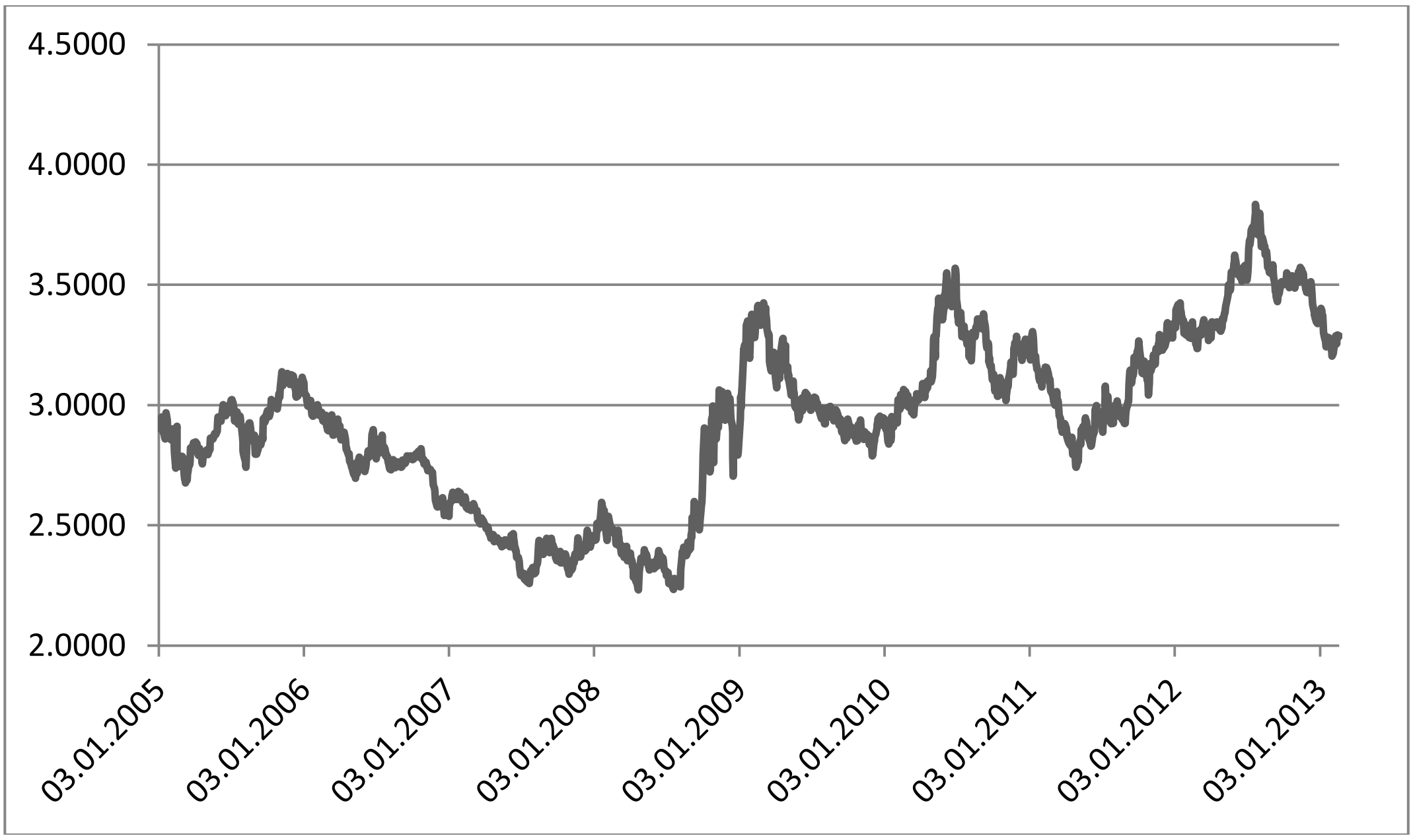

Figure 2. The exchange rate USD/RON

This transformation normalizes the data and guarantee us the values are in [0,1] interval. The two data series transformed according to the method described above are presented in figures 3 and 4.

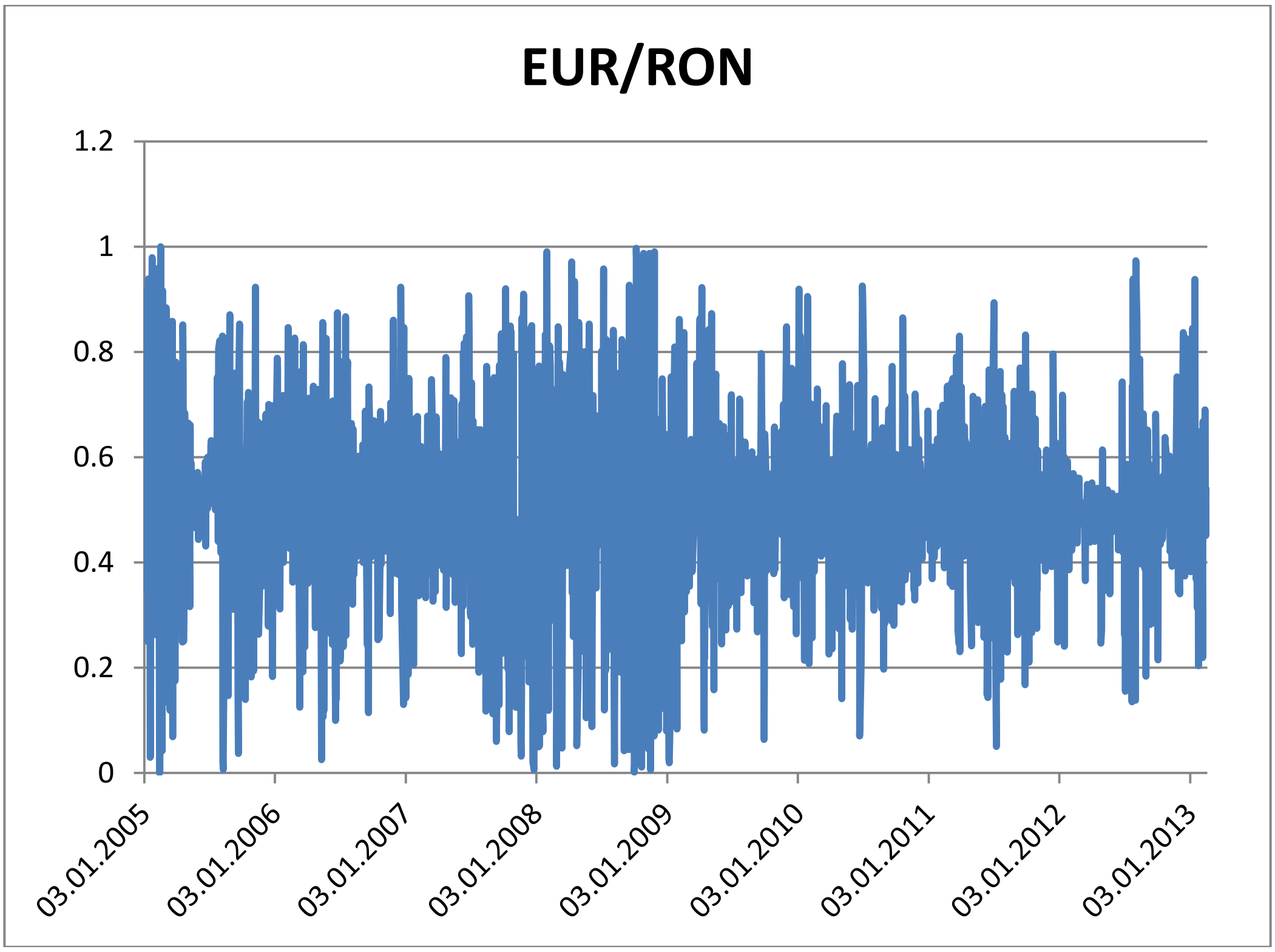

Figure 3. The normalized EUR/RON exchange rate

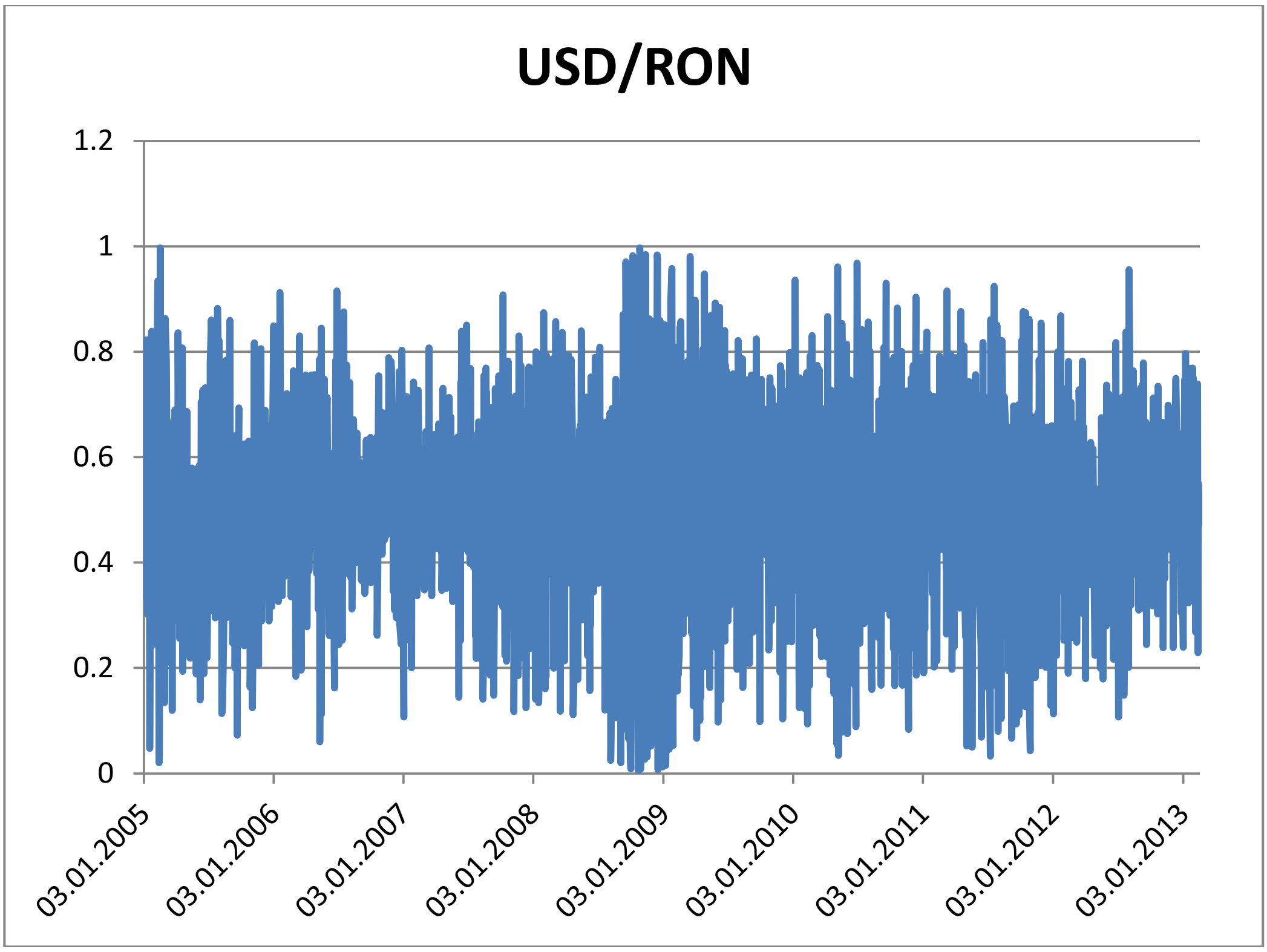

Figure 4. The normalized USD/RON exchange rate

**Results**

First, we built a feed forward neural network with three layers, one input layer, one hidden layer and one output layer. We chose the optimum number of neurons by trial and error: 20 neurons (i.e. we fed into the network the past 20 values of the time series) in the input layer and 40 neurons in the hidden layer. The output layer has one neuron, the value of the time series at t+1 time moment. For the input layer and the output layer we used a linear activation function and *tanh* for the hidden layer. The data sets were spitted in two parts: 80% of the values were used for training and 20% for testing purposes. We tested several training algorithms (Rojas, 1996): backpropagation, resilient propagation and a version of it called RPROP+ which is one of the best performing first-order learning methods for feed forward neural networks (Igel, 2000). The iRPROP+ training algorithm combines information about the sign of the error function derivative which is error surface information with the magnitude of the network error when the decision of reverting an update step is taken. As the error function for our network we used MSE.

The results for the three training algorithms were similar, with an increase in convergence speed of about 25% in the case of iRPROP. As a stopping criteria we imposed a value of 0.1% of the error function.

Second, we tested a recurrent neural network (Jaeger, 2002) with an Elman Simple Recurrent Network architecture trained by Extended Kalman Filter (EKF) method, with the derivatives of the network output computed by Truncated Backpropagation Through Time method (Werbos, 1990), (Zipser, 1995). We used a multistream approach for training the network. If the training data sets are heterogeneous, it is possible that a recency effect may appear – the weights are updated according to the currently presented training data. To circumvent this issue the solution is to scramble the order of the presentation of the data and to use a batch update algorithm. The multistream procedure avoids the apparition of the recency effect by combining the scrambling and batch updates. Multistream EKF divides the training data set into several parts, each subset being a stream. For each *Ns* streams a different instance of the same network is used. A full description of the training procedure (multistream EKF) is beyond the scope of this paper but an interested reader can find this in (Feldkamp, 1998). We used a network with 20 input neurons, 10 fully recurrent hidden neurons and one output neuron. We split the training data in 20 streams randomly chosen each stream with 200 points. The window for truncated backpropagation was 20. With these parameters we obtained an error in forecasting of about 0.01% which is an order of magnitude better than in the case of the feedforward network.

**Conclusions**

Forecasting financial time series is a difficult problem. In this paper paper we compared the performances of different feed forward and recurrent neural networks and training algorithms for predicting the exchange rate EUR/RON and USD/RON. We used data series with daily exchange rates starting from 2005 until 2013 provided by the National Bank of Romania. Before training the networks we applied a preprocessing procedure to the data sets which removes the correlation between data and normalized the data series. Our tests showed that the recurrent network performs better than classical feed forward network in this case.